\documentclass{article}
\usepackage{spconf,amsmath,amssymb,graphicx,booktabs,microtype}
\usepackage[T1]{fontenc}
\usepackage{times}
\usepackage[hidelinks]{hyperref}
\usepackage{balance}
\graphicspath{{figures/}}
\newcommand{\model}{DualTrack}

\title{DualTrack: Synchronized Speech--Gesture Generation\\via Symmetric Coupling of Pretrained Priors}

\name{
Yuanzhuo Hu$^{1}$,
Zehan Liu$^{2}$,
Xiaoyi Qin$^{2}$,
Ming Li$^{1,*}$\thanks{Submitted to ICASSP 2027}
}

\address{
$^{1}$The Chinese University of Hong Kong, Shenzhen, China\\
$^{2}$X Square Robot
}

\begin{document}
\raggedbottom
\maketitle

\begingroup
\renewcommand{\thefootnote}{\fnsymbol{footnote}}
\footnotetext[1]{Corresponding author.
Email: mingli369@cuhk.edu.cn}
\endgroup

\begin{abstract}
Joint speech--gesture synthesis must coordinate two modalities despite limited paired data. Existing approaches often lack bidirectional interaction, have limited language coverage, or simplify body and finger representations. We present DualTrack~\footnote{Code and checkpoints:https://github.com/Yuanzhuo2021/DualTrack}, which couples pretrained speech and motion priors on a shared 12.5\,Hz timeline. Causal adapters exchange previous-packet information, while current-state fusion coordinates the streams before they separately complete sixteen-codebook packets. We evaluate 43 BEAT2 recordings in four languages, with speakers held out from joint training and validation. On the shared English/Spanish inputs, without speech or motion prefixes, DualTrack achieves lower word error rate and full-motion Fr\'echet Gesture Distance, higher beat consistency and speech naturalness than the evaluated GELINA baseline.
\end{abstract}

\begin{keywords}
Speech--gesture synthesis, pretrained motion priors, multimodal generation, text-to-speech generation
\end{keywords}
\section{Introduction}
\label{sec:intro}

Conversational avatars and robots need speech and gestures that express the same intent and follow a shared rhythm. Progress in text-to-speech~\cite{qwen3tts}, text-to-motion~\cite{motiongpt}, and speech-driven gesture synthesis~\cite{emage} has brought this goal closer. In human communication, speech and gestures are closely coordinated~\cite{magi,kendon1997gesture}. Cascaded pipelines, however, capture only one direction of this relationship, letting speech guide motion without feedback~\cite{wang2021integrated}. Unified methods have explored diffusion~\cite{diffttsg}, flow matching~\cite{matchttsg}, and interleaved token prediction~\cite{gelina}. These approaches do not combine bidirectional cross-modal interaction with parallel prediction of temporally aligned packets at each autoregressive step.

Aligned speech--motion corpora offer less data and diversity than large unimodal datasets~\cite{beat,emage,magi}. GELINA~\cite{gelina} uses speech pretraining but learns gesture sequence modeling from paired BEAT2 data; it has limited supported languages, and its evaluated motion representation excludes fingers~\cite{gelina}. MAGI~\cite{magi} transfers unimodal knowledge through synthetic pairs, which may not fully capture natural speech--gesture relationships~\cite{magi}. We directly couple independently pretrained speech and motion generators, retaining multilingual speech capabilities alongside motion dynamics and fine-grained body and finger articulation. Joint adaptation can then use limited paired data primarily to learn cross-modal coordination.

We introduce \model{}, coupling pretrained speech and motion generators on a shared 12.5\,Hz packet clock. Residual adapters exchange previous-packet information bidirectionally, and Tick Fusion coordinates current speech and motion-semantic states before generating motion and speech packets in parallel. DualTrack generates natural, high-quality speech and expressive gestures with fine-grained finger movements, keeping speech and motion closely synchronized.

Our contribution is 
(i) We propose DualTrack, a joint autoregressive framework that connects independently pretrained speech and motion generators through temporally aligned 16-code packets, combining bidirectional interaction with parallel prediction across modalities. (ii) We demonstrate high-quality, multi-language speech and fine-grained body and finger motion generation, with generalization to unseen speakers. (iii) We show that residual adapters and Tick Fusion improve speech–gesture rhythmic alignment, motion diversity, and speech naturalness.

\begin{figure*}[t]
\centering
\includegraphics[width=0.68\textwidth]{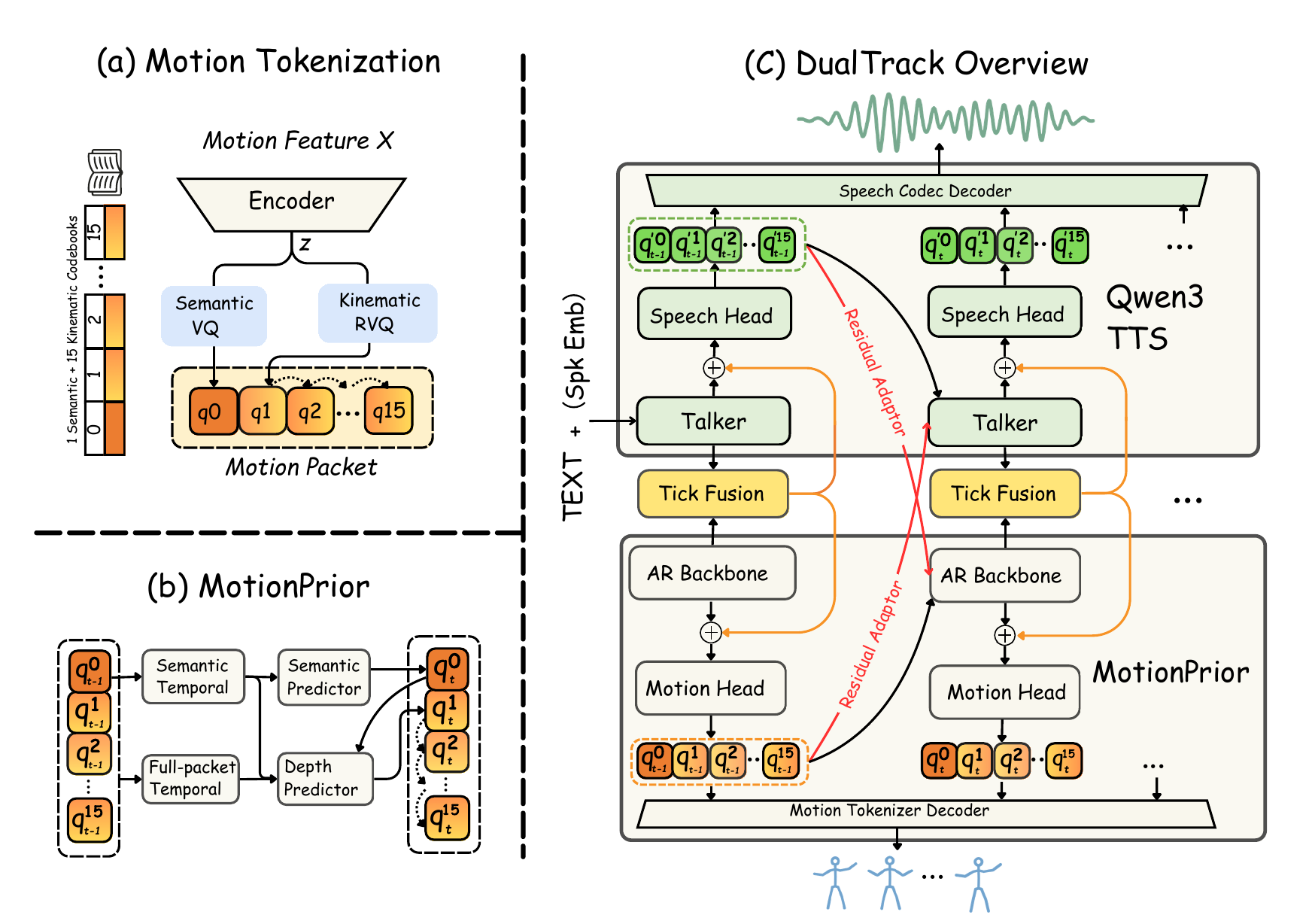}
\caption{\textbf{The overview of DualTrack} (a) Adapt the SeMoCo-based tokenizer with one semantic and fifteen kinematic codebooks. (b) Freeze the tokenizer and pretrain MotionPrior's temporal and depth predictors. (c) Couple MotionPrior and Qwen3-TTS through causal history exchange and Tick Fusion. Each stream predicts sixteen codes per 80\,ms packet.}
\label{fig:architecture}
\end{figure*}

\section{Related Work}
\label{sec:related}
\textbf{Unified speech and gesture synthesis.}
Speech-driven methods generate gestures from available audio~\cite{gesticulator,yoon,diffgesture,emage}; joint synthesis additionally generates speech.
Diff-TTSG~\cite{diffttsg} uses separate diffusion pathways, whereas Match-TTSG~\cite{matchttsg} models concatenated speech and motion features with flow matching~\cite{diffttsg,matchttsg}. GELINA combines speech pretraining with interleaved speech--gesture tokens; its autoregressive backbone retains one motion quantization level and uses a flow-matching decoder to recover detail~\cite{gelina}. DualTrack couples two independently pretrained temporal generators through their states and histories, with neither current packet serving as a prerequisite for the other.

\textbf{Hierarchical discrete priors.}
T2M-GPT~\cite{t2mgpt}, MotionGPT~\cite{motiongpt}, and MoMask~\cite{momask} use discrete motion codes; SeMoCo assigns separate codebooks to semantics and kinematics~\cite{semoco}. Audio codecs provide multicodebook representations~\cite{encodec}, while Moshi and Qwen3-TTS separate temporal prediction from codebook completion~\cite{moshi,qwen3tts}. DualTrack retains a temporal model and depth predictor for each modality and learns residual connections between the two streams.

\section{Methods}
\label{sec:method}

\subsection{Data Preparation}
\label{sec:data_preparation}
We use release910k~\cite{release910k} and BONES-SEED~\cite{bones_seed} for large-scale human motion and diverse motion-captured activities, together with BEAT2~\cite{emage}, which provides paired speech and motion across multiple languages. All motions are standardized to the 77-joint SOMA skeleton and represented as 499-dimensional UMR features at 25 fps. For BEAT2, we retarget the original SMPL-X motion sequences to SOMA, resample them to 25 fps, and extract the corresponding UMR499 features. We follow the dataset-specific partitions for release910k and BONES-SEED. To limit speaker-specific overlap in both speech and motion, we partition BEAT2 into nineteen training, three validation, and three test speakers, with no speaker shared across subsets\footnote{Validation speakers: 6\_carla, 9\_miranda, and 23\_hailing.
Test speakers: 13\_lu, 15\_carlos, and 25\_goto.
Train speakers: the remaining 19 speakers}. All recordings from the same speaker remain in one subset, and temporal windows are extracted only after partitioning.

\subsection{SeMoCo Tokenizer}
We adapt SeMoCo~\cite{semoco} to 499-dimensional UMR499 body
and finger motion at 25~Hz. A shared convolutional encoder
with stride two maps normalized motion to a 512-dimensional
latent $z$. Two parallel branches quantize semantic and
kinematic information, respectively:
\begin{equation}
    \hat{z}
    = \mathcal{Q}_{\mathrm{sem}}(z)
    + \mathcal{Q}_{\mathrm{kin}}(z),
    \label{eq:motion_tokenizer}
\end{equation}
where each branch includes its input and output projections.
The semantic branch uses a single vector quantizer, while
the kinematic branch uses a 15-level residual quantizer,
with 1,024 entries per codebook. Residual subtraction is
confined to the kinematic branch. The summed embeddings are decoded into motion, while the
quantization indices form a motion packet
$m_t=(q_t^0,q_t^1,q_t^2\ldots,q_t^{15})$ at 12.5~Hz.
Here, $q_t^0$ is the semantic code and
$q_t^1,q_t^2,\ldots,q_t^{15}$ are the 15 kinematic codes.

We first adapt the tokenizer on BONES-SEED and BEAT2,
then continue training with release910k added, using
64-frame windows sampled equally from each dataset.
Training combines motion reconstruction, kinematic regularization, and commitment
losses. We retain the checkpoint with the
lowest validation reconstruction MSE and freeze the
tokenizer and normalization statistics.

\subsection{MotionPrior}

We introduce MotionPrior, an autoregressive model over discrete
motion packets produced by the frozen tokenizer. A two-layer
semantic transformer encodes past semantic codes, while a
four-layer full-history transformer encodes complete past
packets, both using up to 150 packets of history. Their output
states, $h_t^m$ and $h_t^f$, are combined as
\begin{equation}
v_t = \operatorname{RMSNorm}\!\left(h_t^f + W_s h_t^m\right),
\label{eq:motion_context_fusion}
\end{equation}
where $W_s$ is a learned linear projection.
At each time step, a factorized semantic predictor first
decides whether to retain the previous semantic code or
switch to a different code. If a switch is selected, it
predicts a new code $q_t^0$ from the remaining entries.
A three-layer Depth GPT then generates the 15 kinematic
codes $q_t^1,q_t^2,\ldots,q_t^{15}$ sequentially, with each prediction conditioned on
$v_t$ and all preceding codes in the current packet,
including the semantic code. All transformers have a
hidden width of 768.

We pretrain MotionPrior on motion-only sequences from the training splits of release910k, BONES-SEED, and BEAT2 described in Section~\ref{sec:data_preparation}. We sample these datasets with probabilities of 87.5\%, 6.25\%, and 6.25\%, respectively, for 761,853 updates to learn a general motion prior through next-packet prediction. We then adapt the model to the BEAT2 motion distribution for an additional 100,000 updates, linearly increasing the BEAT2 sampling probability to 80\% and assigning the remaining sampling probability to release910k.

\subsection{DualTrack}

We initialize the speech and motion streams from Qwen3-TTS~\footnote{Qwen/Qwen3-TTS-12Hz-0.6B-Base}~\cite{qwen3tts}
and MotionPrior. Given text and an optional speaker
embedding $c$, the streams generate speech packets
$s_t=(q_t^{\prime 0},\ldots,q_t^{\prime 15})$ and motion
packets $m_t=(q_t^0,\ldots,q_t^{15})$ at a shared rate of
12.5~Hz. The packets are decoded into audio and motion at
their respective sampling rates.

Residual adapters inject the previous motion packet into
the speech input and the previous speech packet into both
motion temporal inputs. These cross-stream history inputs
are zero at BOS. Text information reaches the motion stream
through the speech stream. Tick Fusion then couples the
current speech state $h_t^s$ and motion-semantic state $h_t^m$:
\begin{equation}
\begin{aligned}
r_t &= \operatorname{SiLU}\!\left(
W_f[\operatorname{LN}(h_t^s);
    \operatorname{LN}(h_t^m)] + b_f
\right), \\
\tilde{h}_t^s &= h_t^s + W_s^{\mathrm{out}}r_t,
\quad
\tilde{h}_t^m = h_t^m + W_m^{\mathrm{out}}r_t.
\end{aligned}
\label{eq:tick_fusion}
\end{equation}
All adapter and fusion output projections are initialized
to zero. The fused states drive the packet predictors.
The motion full-history state bypasses Tick Fusion and
is combined with $\tilde{h}_t^m$ to condition the motion
depth predictor. Neither stream conditions on the other
stream's current output codes, allowing parallel packet
prediction across modalities. Each stream completes its
own packet autoregressively, and both completed packets
provide context for the next tick.

Training combines weighted speech and motion token
cross-entropies with speech replay and distillation.
Only the coupling modules are trained for the first
400 updates. Subsequent updates train the speech-to-motion
adapter, the motion output projection of Tick Fusion,
and the top two blocks and output normalization layers
of each motion temporal transformer. The motion-to-speech
adapter and the shared and speech-facing components of
Tick Fusion are then frozen.
After warm-up, half of the paired examples use generated
motion histories spanning up to 8 ticks initially and
16 ticks later, while retaining recorded speech histories
and ground-truth targets. The pretrained speech model,
motion packet predictors, motion embeddings, and codecs
remain frozen. 

\begin{table*}[t]
\centering
\caption{Speech and gesture generation of Gelina and DualTrack on BEAT2 test subsets.}
\label{tab:dualtrack_multilingual}
\small
\setlength{\tabcolsep}{3pt}
\renewcommand{\arraystretch}{0.97}
\begin{tabular}{llcrrrrrrr}
\toprule
& & & \multicolumn{5}{c}{Gesture} & \multicolumn{2}{c}{Speech} \\
\cmidrule(lr){4-8}\cmidrule(lr){9-10}
Language & Model & $N$
& FGD Full $\downarrow$
& FGD Body $\downarrow$
& BC $\uparrow$
& Div-Body
& Div-Full
& \shortstack{WER/CER(\%) $\downarrow$}
& NMOS $\uparrow$ \\
\midrule
English & Ground truth & 22
& 0.000 & 0.000 & 0.442 & 4.671 & 16.085 & 39.34 & 2.972 \\
& Gelina & 22
& 7.513 & 1.952 & 0.341 & 2.651 & 2.651 & 18.09 & 2.709 \\
& DualTrack & 22
& \underline{7.346} (5.436)
& \underline{1.586} (0.882)
& \textbf{0.606}
& \underline{3.497}
& \underline{10.277}
& \textbf{2.89}
& \textbf{3.847} \\
& DualTrack$^{\dagger}$ & 22
& \textbf{6.889} (5.436)
& \textbf{1.573} (0.882)
& \underline{0.567}
& \textbf{3.570}
& \textbf{11.499}
& \underline{3.45}
& \underline{3.830} \\
\midrule
Chinese & Ground truth & 9
& 0.000 & 0.000 & 0.526 & 4.668 & 16.781 & 5.91 & 2.771 \\
& Gelina & 0/9
& -- & -- & -- & -- & -- & -- & -- \\
& DualTrack & 9
& \underline{9.446} (5.688)
& \underline{2.139} (0.654)
& \underline{0.625}
& \textbf{3.787}
& \textbf{10.502}
& \underline{2.10}
& \underline{3.932} \\
& DualTrack$^{\dagger}$ & 9
& \textbf{8.969} (5.688)
& \textbf{1.924} (0.654)
& \textbf{0.626}
& \underline{3.303}
& \underline{10.273}
& \textbf{1.66}
& \textbf{4.229} \\
\midrule
Japanese & Ground truth & 6
& 0.000 & 0.000 & 0.309 & 3.191 & 13.662 & 10.06 & 3.769 \\
& Gelina & 0/6
& -- & -- & -- & -- & -- & -- & -- \\
& DualTrack & 6
& \underline{7.874} (4.801)
& \textbf{2.103} (0.999)
& \textbf{0.553}
& \underline{3.982}
& \textbf{11.732}
& \textbf{4.56}
& \underline{3.954} \\
& DualTrack$^{\dagger}$ & 6
& \textbf{7.832} (4.801)
& \underline{2.170} (0.999)
& \underline{0.474}
& \textbf{2.956}
& \underline{9.151}
& \underline{4.73}
& \textbf{3.987} \\
\midrule
Spanish & Ground truth & 6
& 0.000 & 0.000 & 0.495 & 5.220 & 17.106 & 10.10 & 2.818 \\
& Gelina & 6
& 9.850 & 3.451 & 0.416 & 2.626 & 2.626
& \underline{215.09} & 2.757 \\
& DualTrack & 6
& \underline{9.539} (5.647)
& \textbf{3.222} (1.391)
& \textbf{0.638}
& \textbf{3.908}
& \textbf{10.567}
& \textbf{0.00}
& \textbf{4.307} \\
& DualTrack$^{\dagger}$ & 6
& \textbf{8.560} (5.647)
& \underline{3.247} (1.391)
& \underline{0.629}
& \underline{3.287}
& \underline{9.456}
& \textbf{0.00}
& \underline{4.081} \\
\bottomrule
\end{tabular}
\vspace{2pt}
\begin{minipage}{\textwidth}
\small
$\dagger$ adds a speaker embedding; unmarked DualTrack and
Gelina are text-only.
Parenthesized FGD values report motion reconstruction
through the frozen tokenizer on the same test subset
and are excluded from ranking.
Bold/underline: best/second-best generated scores per language;
ties share ranks.
$N$: recordings; --: unavailable/unreported.
WER: English/Spanish; CER: Chinese/Japanese.BC represents for beat consistency.
\end{minipage}
\end{table*}

\section{Experiments}
\label{sec:experiments}
\subsection{Evaluation Protocol}

We evaluate the intersection of our held-out test speakers
and the official BEAT2 test set, comprising 43 recordings:
22 English, 9 Chinese, 6 Japanese, and 6 Spanish.
Each reference window lasts 7.68~s, and generation starts
from native BOS without a motion prefix. DualTrack and
GELINA use text only; DualTrack$^\dagger$ additionally uses
a speaker embedding from a five-second reference taken
from another recording of the same speaker. Both DualTrack
variants use the same checkpoint across languages.
Direct comparisons with GELINA use the 28 English/Spanish
inputs supported by its text tokenizer. Generation uses
FP32 with TF32 disabled, batch size one, shared per-recording
seeds.

\textbf{Metrics.}
We compute full/body FGD~\cite{fid} with the frozen GELINA/EMAGE encoder~\cite{gelina,emage}, centered 240-frame windows, and repeat-last padding. Body FGD zeros finger rotation-6D features~\cite{rot6d}. DualTrack motion follows UMR499 $\rightarrow$ SOMA~\cite{soma} $\rightarrow$ SMPL-X~\cite{smplx} through official inverse fitting; representation conversion, retargeting, and resampling introduce residual error. Body/full L1 diversity measures temporal motion variation; values closer to ground truth are preferred. Beat Consistency is computed over the full overlapping duration of the audio and generated motion. We score Whisper-large-v3~\cite{whisper} transcripts with unclipped WER for English/Spanish and CER for Chinese/Japanese. NISQA-TTS~\cite{nisqa} predicts speech naturalness (NMOS). 

\subsection{Evaluation}
On the shared English/Spanish inputs (Table 1), DualTrack†
lowers English WER from GELINA’s 18.09\% to 3.45\% and
raises NMOS from 2.709 to 3.830. Full FGD falls from 7.513
to 6.889; full diversity moves from 2.651 to 11.499 toward GT
16.085. On six Spanish recordings, WER falls from 215.09\%
to 0.00\%, and Full/Body FGD decreases from 9.850/3.451 to
8.560/3.247. Text-only DualTrack gives English WER 2.89\%
and Full FGD 7.345. Chinese/Japanese CER is 2.10\%/4.56\%
without a speaker embedding and 1.66\%/4.73\% with one, with-
out a supported GELINA comparison. Full-motion scores also
reflect GELINA’s lack of finger articulation [9].

\textbf{Ablation Study.}
Table~\ref{tab:ablation_a_d} compares the full model with
the variant disabling both history exchange and current-tick
fusion. The full model achieves lower full-body FGD, higher beat
consistency, and higher predicted speech naturalness.
Its full-body motion diversity is also closer to the matched
ground truth, supporting the joint contribution of these
components to motion generation and speech naturalness.

\begin{table}[t]
\centering
\caption{Ablation results on the complete
test split (136 recordings). Div.\ denotes full-body motion diversity. BC
 denotes beat consistency}
\label{tab:ablation_a_d}
\small
\begin{tabular}{lcccc}
  \toprule
  Model & Full FGD$\downarrow$ & BC$\uparrow$
        & Div. & NMOS$\uparrow$ \\
  \midrule
  Full DualTrack & \textbf{5.7175} & \textbf{0.5653}
    & \textbf{10.582} & \textbf{3.885} \\
  w/o residual+fusion & 5.9637 & 0.5394 & 10.015 & 3.619 \\
  \bottomrule
\end{tabular}
\end{table}

\begin{figure}[t]
  \centering
  \includegraphics[width=\columnwidth]{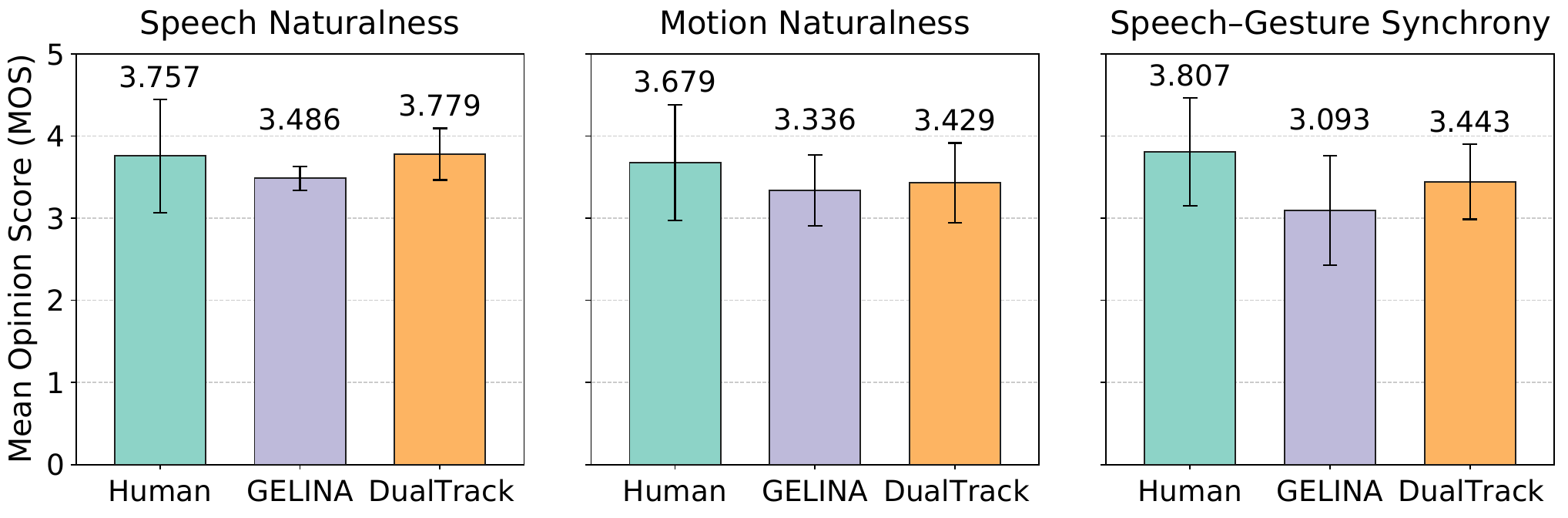}
  \caption{Subjective evaluation results. Error bars indicate
  $\pm 1$ standard deviation across participant-level mean scores.}
  \label{fig:user_study}
\end{figure}

\textbf{User Study.}
We conducted a subjective evaluation of speech naturalness, motion
naturalness, and speech--gesture synchrony using audio-only,
motion-only, and audio-visual stimuli, respectively (Figure~\ref{fig:user_study}).
A total of 15 Participants rated Human, GELINA, and DualTrack outputs on a
five-point scale, with model identities hidden and candidate order
randomized. For each task, we retained five participants who completed
all 28 shared samples (22 English + 6 Spanish). DualTrack outperforms GELINA in all three tasks.

\textbf{Conversion-matched reference.}
A separate diagnostic passes unquantized GT UMR499 features through the same SOMA-to-SMPL-X conversion as archived FP32 DualTrack$^{\dagger}$ outputs. On the 28 English/Spanish recordings, pooled Full/Body FGD is 2.956/1.186 against this converted GT, versus 6.693/1.693 against native GT. Converted GT itself has Full/Body FGD 3.750/0.602 against native GT. Despite the discrepancies introduced by conversion, DualTrack still achieves strong performance against native ground truth.

\section{Conclusion and Future Improvements}
DualTrack couples pretrained speech and motion priors through bidirectional history exchange and current-state fusion. On shared English/Spanish inputs, it achieves higher predicted speech naturalness, lower FGD for full-body motion including fingers, and stronger speech–gesture beat alignment than GELINA, supporting more coordinated speech and gesture generation. Its causal packet interface provides a natural basis for streaming generation. Future work will build on this structure with incremental decoding and conversions, aiming to enable low-latency, real-time speech and gesture synthesis.

\clearpage

\section{Acknowledgments}
The authors used OpenAI GPT-6-Astra to improve the grammar and wording. LLM was used solely for language-related improvements. All AI-assisted content was reviewed and approved by authors. The authors take full responsibility for the final manuscript.

\begingroup
\small
\balance
\bibliographystyle{IEEEbib}
\bibliography{references}

@inproceedings{gelina,
  author = {Guichoux, T{\'e}o and Lemerle, Th{\'e}odor and Mehta, Shivam and Beskow, Jonas and Henter, Gustav Eje and Soulier, Laure and others},
  title = {{Gelina: Unified Speech and Gesture Synthesis via Interleaved Token Prediction}},
  booktitle = {Proc. IEEE ICASSP},
  year = {2026},
  pages = {16122--16126},
  doi = {10.1109/ICASSP55912.2026.11464562},
  url = {https://arxiv.org/abs/2510.12834}
}

@article{qwen3tts,
  author = {Hu, Hangrui and Zhu, Xinfa and He, Ting and Guo, Dake and Zhang, Bin and Wang, Xiong and others},
  title = {{Qwen3-TTS Technical Report}},
  journal = {arXiv preprint arXiv:2601.15621},
  year = {2026},
  url = {https://arxiv.org/abs/2601.15621}
}

@article{semoco,
  author = {Huang, Tianlv and Guo, Hetian and Cai, Ziyi and Wang, Song and Zhang, Yanping and Fan, Zipei and others},
  title = {{SeMoCo: A Semantic-First Motion Codec for Motion Language Modeling}},
  journal = {arXiv preprint arXiv:2608.24334},
  year = {2026},
  url = {https://arxiv.org/abs/2608.24334}
}

@inproceedings{emage,
  author = {Liu, Haiyang and Zhu, Zihao and Becherini, Giorgio
               and Peng, Yichen and Su, Mingyang and Zhou, You
               and Zhe, Xuefei and Iwamoto, Naoya and Zheng, Bo
               and Black, Michael J.},
  title = {{EMAGE: Towards Unified Holistic Co-Speech Gesture Generation via Expressive Masked Audio Gesture Modeling}},
  booktitle = {Proc. IEEE/CVF CVPR},
  year = {2024},
  pages = {1144--1154},
  url = {https://arxiv.org/abs/2401.00374}
}

@inproceedings{motiongpt,
  author = {Jiang, Biao and Chen, Xin and Liu, Wen and Yu, Jingyi and Yu, Gang and Chen, Tao},
  title = {{MotionGPT: Human Motion as a Foreign Language}},
  booktitle = {Adv. Neural Inf. Process. Syst.},
  year = {2023},
  url = {https://arxiv.org/abs/2306.14795}
}

@inproceedings{matchttsg,
  author = {Mehta, Shivam and Tu, Ruibo and Alexanderson, Simon and Beskow, Jonas and Sz{\'e}kely, {\'E}va and Henter, Gustav Eje},
  title = {{Unified speech and gesture synthesis using flow matching}},
  booktitle = {Proc. IEEE ICASSP},
  year = {2024},
  url = {https://arxiv.org/abs/2310.05181}
}

@inproceedings{diffttsg,
  author = {Mehta, Shivam and Wang, Siyang and Alexanderson, Simon and Beskow, Jonas and Sz{\'e}kely, {\'E}va and Henter, Gustav Eje},
  title = {{Diff-TTSG: Denoising probabilistic integrated speech and gesture synthesis}},
  booktitle = {Proc. ISCA SSW},
  pages = {150--156},
  doi = {10.21437/SSW.2023-24},
  year = {2023},
  url = {https://arxiv.org/abs/2306.09417}
}

@inproceedings{magi,
  author = {Mehta, Shivam and Deichler, Anna and O'Regan, Jim and Mo{\"e}ll, Birger and Beskow, Jonas and Henter, Gustav Eje and others},
  title = {{Fake It to Make It: Using Synthetic Data to Remedy the Data Shortage in Joint Multimodal Speech-and-Gesture Synthesis}},
  booktitle = {Proc. IEEE/CVF CVPR Workshops},
  year = {2024},
  pages = {1952--1964},
  url = {https://openaccess.thecvf.com/content/CVPR2024W/HuMoGen/html/Mehta_Fake_It_to_Make_It_Using_Synthetic_Data_to_Remedy_CVPRW_2024_paper.html}
}

@inproceedings{momask,
  author = {Guo, Chuan and Mu, Yuxuan and Javed, Muhammad Gohar and Wang, Sen and Cheng, Li},
  title = {{MoMask: Generative Masked Modeling of 3D Human Motions}},
  booktitle = {Proc. IEEE/CVF CVPR},
  year = {2024},
  pages = {1900--1910},
  url = {https://arxiv.org/abs/2312.00063}
}

@article{moshi,
  author = {D{\'e}fossez, Alexandre and Mazar{\'e}, Laurent and Orsini, Manu and Royer, Am{\'e}lie and P{\'e}rez, Patrick and J{\'e}gou, Herv{\'e} and others},
  title = {{Moshi: a speech-text foundation model for real-time dialogue}},
  journal = {arXiv preprint arXiv:2410.00037},
  year = {2024},
  url = {https://arxiv.org/abs/2410.00037}
}

@article{yoon,
  author = {Yoon, Youngwoo and Cha, Bok and Lee, Joo-Haeng and Jang, Minsu and Lee, Jaeyeon and Kim, Jaehong and others},
  title = {{Speech Gesture Generation from the Trimodal Context of Text, Audio, and Speaker Identity}},
  journal = {ACM Transactions on Graphics},
  year = {2020},
  volume = {39},
  number = {6},
  url = {https://arxiv.org/abs/2009.02119}
}

@inproceedings{smplx,
  author = {Pavlakos, Georgios and Choutas, Vasileios and Ghorbani, Nima and Bolkart, Timo and Osman, Ahmed A. A. and Tzionas, Dimitrios and others},
  title = {{Expressive Body Capture: 3D Hands, Face, and Body from a Single Image}},
  booktitle = {Proc. IEEE/CVF CVPR},
  year = {2019},
  url = {https://arxiv.org/abs/1904.05866}
}

@inproceedings{rot6d,
  author = {Zhou, Yi and Barnes, Connelly and Lu, Jingwan and Yang, Jimei and Li, Hao},
  title = {{On the Continuity of Rotation Representations in Neural Networks}},
  booktitle = {Proc. IEEE/CVF CVPR},
  year = {2019},
  url = {https://arxiv.org/abs/1812.07035}
}

@inproceedings{fid,
  author = {Heusel, Martin and Ramsauer, Hubert and Unterthiner, Thomas and Nessler, Bernhard and Hochreiter, Sepp},
  title = {{GANs Trained by a Two Time-Scale Update Rule Converge to a Local Nash Equilibrium}},
  booktitle = {Adv. Neural Inf. Process. Syst.},
  year = {2017},
  url = {https://arxiv.org/abs/1706.08500}
}

@inproceedings{gesticulator,
  author = {Kucherenko, Taras and Jonell, Patrik and van Waveren, Sanne and Henter, Gustav Eje and Alexanderson, Simon and Leite, Iolanda and others},
  title = {{Gesticulator: A Framework for Semantically-Aware Speech-Driven Gesture Generation}},
  booktitle = {Proc. ACM ICMI},
  year = {2020},
  doi = {10.1145/3382507.3418815},
  url = {https://arxiv.org/abs/2001.09326}
}

@inproceedings{diffgesture,
  author = {Zhu, Lingting and Liu, Xian and Liu, Xuanyu and Qian, Rui and Liu, Ziwei and Yu, Lequan},
  title = {{Taming Diffusion Models for Audio-Driven Co-Speech Gesture Generation}},
  booktitle = {Proc. IEEE/CVF CVPR},
  year = {2023},
  pages = {10544--10553},
  url = {https://openaccess.thecvf.com/content/CVPR2023/html/Zhu_Taming_Diffusion_Models_for_Audio-Driven_Co-Speech_Gesture_Generation_CVPR_2023_paper.html}
}

@inproceedings{beat,
  author = {Liu, Haiyang and Zhu, Zihao and Iwamoto, Naoya and Peng, Yichen and Li, Zhengqing and Zhou, You and others},
  title = {{BEAT: A Large-Scale Semantic and Emotional Multi-Modal Dataset for Conversational Gestures Synthesis}},
  booktitle = {Proc. ECCV},
  year = {2022},
  url = {https://arxiv.org/abs/2203.05297}
}

@inproceedings{t2mgpt,
  author = {Zhang, Jianrong and Zhang, Yangsong and Cun, Xiaodong and Zhang, Yong and Zhao, Hongwei and Lu, Hongtao and others},
  title = {{Generating Human Motion from Textual Descriptions with Discrete Representations}},
  booktitle = {Proc. IEEE/CVF CVPR},
  year = {2023},
  pages = {14730--14740},
  url = {https://openaccess.thecvf.com/content/CVPR2023/html/Zhang_Generating_Human_Motion_From_Textual_Descriptions_With_Discrete_Representations_CVPR_2023_paper.html}
}

@article{encodec,
  author = {D{\'e}fossez, Alexandre and Copet, Jade and Synnaeve, Gabriel and Adi, Yossi},
  title = {{High Fidelity Neural Audio Compression}},
  journal = {arXiv preprint arXiv:2210.13438},
  year = {2022},
  url = {https://arxiv.org/abs/2210.13438}
}

@inproceedings{whisper,
  author = {Alec Radford and Jong Wook Kim and Tao Xu and Greg Brockman and Christine McLeavey and Ilya Sutskever},
  title = {Robust Speech Recognition via Large-Scale Weak Supervision},
  booktitle = {Proc. ICML},
  year = {2023},
  pages = {28492--28518},
  url = {https://proceedings.mlr.press/v202/radford23a.html}
}

@inproceedings{nisqa,
  author = {Gabriel Mittag and Babak Naderi and Assmaa Chehadi and Sebastian M{\"o}ller},
  title = {{NISQA}: A Deep {CNN}-Self-Attention Model for Multidimensional Speech Quality Prediction with Crowdsourced Datasets},
  booktitle = {Proc. Interspeech},
  year = {2021},
  pages = {2127--2131},
  doi = {10.21437/Interspeech.2021-299}
}

@article{soma,
  author = {Saito, Jun and Li, Jiefeng and de Ruyter, Michael and Guerrero, Miguel and Lim, Edy and Hassani, Ehsan and others},
  title = {{SOMA: Unifying Parametric Human Body Models}},
  journal = {arXiv preprint arXiv:2603.16858},
  year = {2026},
  url = {https://arxiv.org/abs/2603.16858}
}

@article{kendon1997gesture,
  title={Gesture},
  author={Kendon, Adam},
  journal={Annual review of anthropology},
  volume={26},
  number={1},
  pages={109--128},
  year={1997},
  publisher={Annual Reviews 4139 El Camino Way, PO Box 10139, Palo Alto, CA 94303-0139, USA}
}

@inproceedings{wang2021integrated,
  author    = {Wang, Siyang and Alexanderson, Simon and
               Gustafson, Joakim and Beskow, Jonas and
               Henter, Gustav Eje and Sz{\'e}kely, {\'E}va},
  title     = {Integrated Speech and Gesture Synthesis},
  booktitle = {Proc. ACM ICMI},
  year      = {2021},
  pages     = {177--185},
  publisher = {ACM},
  doi       = {10.1145/3462244.3479914}
}

@misc{release910k,
  title        = {{SOMA UMR Release v260717 T2M 910k (realigned)}},
  howpublished = {Hugging Face dataset repository.
                  \url{https://huggingface.co/datasets/poisonousID/release-910k-new}},
  note         = {Accessed: September 22, 2026}
}

@misc{bones_seed,
  author       = {{Bones Studio}},
  title        = {{BONES-SEED}: Skeletal Everyday Embodiment Dataset},
  howpublished = {Hugging Face dataset repository.
                  \url{https://huggingface.co/datasets/bones-studio/seed}},
  note         = {Accessed: September 22, 2026}
}
\endgroup
\end{document}